\documentclass[conference]{IEEEtran}
\IEEEoverridecommandlockouts
\usepackage{cite}
\usepackage{amsmath,amssymb,amsfonts}
\usepackage{algorithmic}
\usepackage{graphicx}
\usepackage{textcomp}
\usepackage{xcolor}

\usepackage{booktabs}
\usepackage{subcaption}

\def\BibTeX{{\rm B\kern-.05em{\sc i\kern-.025em b}\kern-.08em
    T\kern-.1667em\lower.7ex\hbox{E}\kern-.125emX}}
\begin{document}

\title{An Exploratory Assessment of LLM's Potential Toward  Flight Trajectory Reconstruction Analysis\\
}

\author{\IEEEauthorblockN{Qilei Zhang}
\IEEEauthorblockA{\textit{School of Aviation and Transportation Technology} \\
\textit{Purdue University}\\
West Lafayette, USA \\
zhan3599@purdue.edu}
\and
\IEEEauthorblockN{John H. Mott}
\IEEEauthorblockA{\textit{School of Aviation and Transportation Technology} \\
\textit{Purdue University}\\
West Lafayette, USA \\
jhmott@purdue.edu}
}

\maketitle

\begin{abstract}
Large Language Models (LLMs) hold transformative potential in aviation, particularly in reconstructing flight trajectories. This paper investigates this potential, grounded in the notion that LLMs excel at processing sequential data and deciphering complex data structures. Utilizing the LLaMA 2 model, a pre-trained open-source LLM, the study focuses on reconstructing flight trajectories using Automatic Dependent Surveillance-Broadcast (ADS-B) data with irregularities inherent in real-world scenarios. The findings demonstrate the model's proficiency in filtering noise and estimating both linear and curved flight trajectories. However, the analysis also reveals challenges in managing longer data sequences, which may be attributed to the token length limitations of LLM models. The study's insights underscore the promise of LLMs in flight trajectory reconstruction and open new avenues for their broader application across the aviation and transportation sectors.
\end{abstract}


\section{Introduction}

The Large Language Models (LLM) are increasingly regarded as a promising approach towards achieving Artificial General Intelligence (AGI) since these models demonstrate an aptitude to handle a diverse range of capabilities akin to those of humans~\cite{xi2023rise}. Particularly, the LLMs have demonstrated great potential in the domain of processing sequential data, notably in natural language processing and interpreting complex data structures, including images and videos. This capability has led to the hypothesis that LLMs could effectively learn and discern patterns in flight trajectories, thereby enabling the reconstruction of flight trajectories that were originally recorded from Automatic Dependent Surveillance-Broadcast (ADS-B) system from the discontinuous, inaccurate, or missing situations~\cite{zhang2021development}.

The ADS-B system is a pivotal surveillance system that broadcasts the aircraft's position, velocity, and other information to the ground station and surrounding aircraft~\cite{ADSB2021FAQ}. Despite its widespread implementation, the data recorded from ADS-B systems are susceptible to various irregularities due to multiple reasons, including but not limited to antennas' blind zones, signal jamming, and multipath interference~\cite{mott2017estimation}. Such deficiencies in ADS-B data present considerable challenges in accurately reconstructing flight trajectories, an endeavor of critical importance in aviation safety~\cite{shi2018lstm}. Specifically, effective trajectory reconstruction aids in identifying potential conflict points between aircraft, detecting abnormal flight behaviors, and enhancing air traffic management systems.

The introduction of one-shot, few-shot, and fine-tuning learning techniques in the realm of LLMs has enabled a new paradigm of learning that requires significantly less training data than conventional machine learning methodologies~\cite{hegselmann2023tabllm}. Moreover, the proficiency of LLMs in encoding sequential data into numerical sequences and their capability in next-token prediction for text generation underscore their potential in processing sequential data~\cite{gruver2023large}. Therefore, this new paradigm of learning is considerably promising for flight trajectory reconstruction.

\section{Literature Review}
The primary challenge addressed in this study is to assess the potential of LLMs in flight trajectory reconstruction using flight data recorded from ADS-B systems with irregularities. 

\subsection{Flight Trajectory Reconstruction}

The conventional methodologies for flight trajectory reconstruction can be categorized into three groups: (1) physical motion functions based on aerodynamics or aircraft performance, (2) linear quadratic estimation (Kalman filtering), and (3) data-driven machine learning models~\cite{ma2020hybrid}.

In the first category, methods utilizing physical motion function derived from the aircraft's aerodynamics, performance characteristics, aircraft intent, flight plans, and performance models have been explored~\cite{yepes2007new, porretta2008performance, thipphavong2013adaptive}. However, these approaches often require extensive parameters and simplifying assumptions, which can compromise accuracy and applicability. The second category, Kalman Filtering, offers efficient real-time processing and accurate trajectory reconstruction~\cite{simon2006optimal}. Its limitation becomes apparent when the aircraft's motion is nonlinear, with unpredictable maneuvers and disturbances in dynamic environments~\cite{dy2021validation}. The third category, data-driven machine learning models, with studies employing boosted regression trees, random forests, and neural networks, have demonstrated promising results in flight trajectory reconstruction~\cite{zeng2020deep, zhang2021development, zhang2022improved}. However, these models require extensive training data with ground truth values, which are often unavailable in real-world scenarios.

\subsection{LLM on Time-Series Data}

The application of LLMs to time series data has gained attention in recent years, as highlighted by several innovative studies. Chang et al.~\cite{chang2023llm4ts} proposed a novel LLM (LLM4TS), a two-stage fine-tuning process for time series forecasting tasks. Their study combined time-series patching with temporal encoding using pre-trained LLMs, offering enhanced forecasting in limited data scenarios. Gruver et al.~\cite{gruver2023large} demonstrated that LLMs, like GPTs and LLaMA, can perform zero-shot time-series forecasting by encoding data as numerical strings. Zhou et al.~\cite{zhou2023one} explored the use of a Frozen Pretrained Transformer (FPT) for time-series forecasting, emphasizing cross-modality applicability. Additionally, the framework proposed by Jin et al.~\cite{jin2023time} reprogramed existing LLMs for time series analysis without altering the original models, outperforming specialized models in various learning scenarios. These studies underline the potential of LLMs in managing and analyzing sequential, multidimensional time-series data effectively.

\section{Methodology}

\subsection{Flight Data Collection and Generation}
This study utilizes BlueSky, an open-source Python package for aircraft operation simulation~\cite{hoekstra2016bluesky}, to generate a synthetic flight dataset. The dataset, designed to mirror the statistical characteristics of ideal, integrity-rich ground truthing flight data, will facilitate both training and evaluation of the proposed model. 

The simulation created diverse flight operations between airports, such as from KLAF to KVPZ, incorporating a range of maneuvers. This diversity in flight operations is achieved by defining a set of waypoints in a rectangular area based on the airports' locations. Within this area, smaller rectangles are created, and their vertices are randomly selected as waypoints for flight trajectories, each assigned with specific altitudes and speeds. Two possible behaviors, flyover and flyby, are simulated in BlueSky, ensuring varied and unique missions for each aircraft initialization.

\begin{figure}[htbp]
    \centering
    \includegraphics[width=0.48\textwidth]{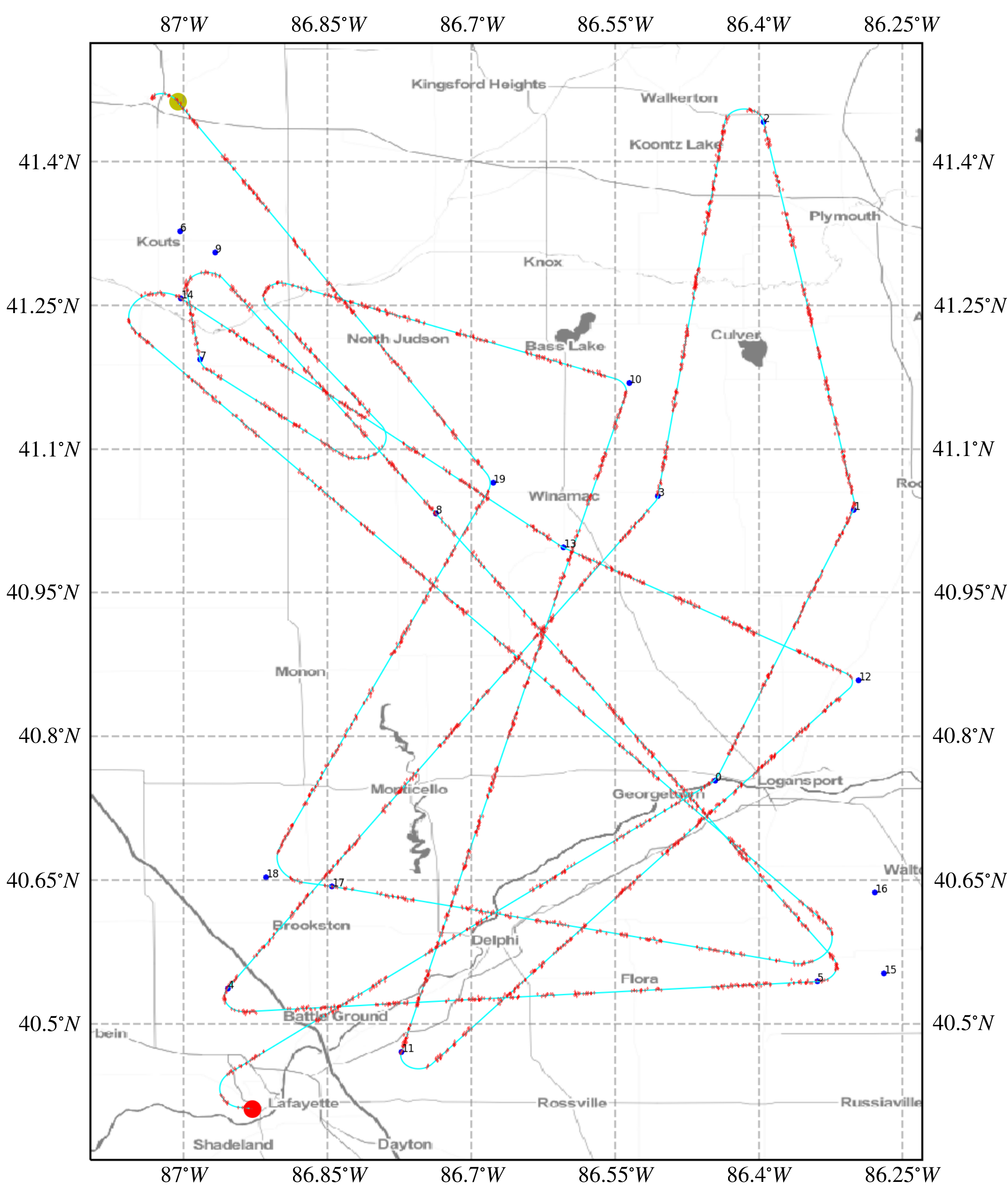}
    \caption{An illustration of a simulated flight trajectory with synthetic ADS-B data departing from KLAF and arriving at KVPZ.}\label{fig:flight_simulation}
\end{figure}

During the simulation, BlueSky will record flight data, including the aircraft's position, velocity, and other information, at a frequency of 1 Hz. The recorded data will be treated as having the same integrity as ground truthing data. To realistically reflect the irregularities in ADS-B data, the study introduced various irregularities into the synthetic flight data, simulating defects and inaccuracies commonly observed in ADS-B data. Consequently, this rendered the synthetic data noisy, characterized by an inconsistent sampling rate and the presence of missing data points. Figure~\ref{fig:flight_simulation} provided a visualization of a generated flight trajectory with simulated ADS-B data, distinguishing the trajectory in cyan-blue and ADS-B data points in red.

\subsection{LLM Model Configuration}

The study adopted the LLaMA 2 model~\cite{touvron2023llama}, a pre-trained LLM model that has demonstrated promising results in time-series forecasting~\cite{gruver2023large}. The LLaMA 2 model offers three model sizes based on the number of parameters, namely 7B, 13B, and 70B. The study will select the 7B model due to its relatively small size and computational efficiency. The model's fine-tuning will utilize the synthetic flight data, processed by two A100 GPUs, each equipped with 40GB memory. 

Considering the LLaMA 2 model's default strategy with the tokenization of numbers into individual digits~\cite{touvron2023llama}, the experiment in the study will forego decimals for floating-point numbers across all attributes, maintaining fixed precision for each. Specifically, inputs will include the common attributes in ADS-B data, namely time, latitude, longitude, altitude, true airspeed, vertical speed, and track angle. Longitude and latitude will have a precision of 0.00001 degrees. Altitude will have a precision of 1 foot. True airspeed will have a precision of 1 knot. Vertical speed will have a precision of 0.1 feet per minute, and track angle will have a precision of 1 degree. Following the recommendation of Gruver et al.~\cite{gruver2023large}, all numbers will be stripped of decimals before feeding into the model's tokenizer. For instance, ``40.12340'' will be converted to ``4012340'' before tokenization. After the trained model's inference, the digits will be restored to the original precision for visualization and analysis.  

Particularly, the experiment setup also incorporates special prompts to guide and inform the LLM in processing flight trajectory data. Each input sequence will commence with an introductory prompt, ``\textit{Determine the curved flight trajectory using these estimated parameters (time, latitude, longitude, altitude, true airspeed, vertical speed, and track angle). Please summarize the precise trajectory considering these inputs: }''. This prompt serves to orient the LLM towards the specific task of flight trajectory reconstruction. Concluding each input sequence, the prompt ``\textit{Summary: }'' will be appended to the end of the input sequence, signaling the end of the input sequence. 

To better facilitate the usage of the computational resources, the study also incorporates the Parameter-Efficient Fine-Tuning (PEFT) techniques, aiming to achieve considerable performance with reduced training time, memory, and data requirements. Specifically, the PEFT in this study was implemented using the Low-rank adaptation of large language models (LoRA) method~\cite{hu2021lora} with the model training configuration. Additionally, the \textit{half-precision floating-point format} has been adopted to reduce memory consumption further. The experimental setup, as outlined in the \textit{llama-recipes} repository~\cite{touvron2023llama}, was followed and adapted for fine-tuning the LLaMA 2 model.

\subsection{Data Preprocessing}

In the research experiment, the dataset was divided into training, validation, and testing subsets in a ratio of 0.8, 0.1, and 0.1, respectively. To align the LLaMA 2 - 7B model's optimal processing capabilities, each input time duration was constrained to 60 seconds, reflecting the model's recommended maximum sequence length of 2048 tokens~\cite{touvron2023llama}. The output is designed to reconstruct the flight trajectory at a five-second interval on integer timestamps, balancing conserving token space and maintaining a clear, observable trajectory representation. The objective is to fine-tune the model with a restricted number of training data while striving to achieve comparable performance. Flight trajectories exhibiting altitude variations of more than 300 feet or cumulative track angle changes of more than 30 degrees were included in the training, validation, and testing datasets. Consequently, the current training, validation, and testing datasets consist of 7281, 911, and 910 flight trajectories, respectively, each spanning a duration of 60 seconds.

\section{Results}

\subsection{Base Model Evaluation}

After preparing and loading the preprocessed dataset, the study first conducted an initial examination of the base LLaMA 2 model prior to any fine-tuning. The input was formatted as suggested in the previous section. Table~\ref{tab:prompt} provides an example of the input prompt.

\begin{table}[htbp]
    \centering
    \caption{An example of an input prompt for the model.}\label{tab:prompt}
    \begin{tabular}{l}
    \toprule
    \texttt{Determine the curved flight trajectory using}\\
    \texttt{these estimated parameters (time, latitude, } \\
    \texttt{longitude, altitude, true airspeed, vertical} \\
    \texttt{speed, and track angle). Please summarize the} \\
    \texttt{precise trajectory considering these inputs:} \\

    \texttt{(967, 4140614, 8692362, 4863, 81, 0, 308),} \\
    \texttt{(1158, 4140473, 8692565, 4895, 81, 0, 308),} \\
    \texttt{(1266, 4140443, 8692432, 4886, 81, 0, 308),} \\
    \texttt{\textit{... continue with other rows ...}} \\
    \texttt{(5747, 4142412, 8696346, 4871, 81, 0, 271)} \\
    \texttt{- - - - - - -} \\
    \texttt{Summary:} \\

    \bottomrule
    \end{tabular}
\end{table}

However, the preliminary evaluation revealed the limitation of the base model in its unmodified state. Rather than generating meaningful outputs, the model tended to repetitively  echo the data part of the input prompt, as the output in Table~\ref{tab:prompt} illustration would be like \texttt{(967, ..., 271)}. A similar outcome was observed when the input data was not stripped of decimals, underscoring the model's inability to comprehend and execute the task of flight trajectory reconstruction in its base state. This observation leads to the necessity of fine-tuning the model to enable it to interpret the task objective  and generate meaningful outputs for flight trajectory reconstruction analysis.

However, the preliminary evaluation revealed the limitation of the base model in its unmodified state. Rather than generating meaningful outputs, the model tended to repetitively  echo the data part of the input prompt, as the output in Table~\ref{tab:prompt} illustration would be like \texttt{(967, $\ldots$, 271)}. A similar outcome was observed when the input data was not stripped of decimals, underscoring the model's inability to comprehend and execute the task of flight trajectory reconstruction in its base state. This observation leads to the necessity of fine-tuning the model to enable it to interpret the task objective  and generate meaningful outputs for flight trajectory reconstruction analysis.

\subsection{Fine-tuning Model Evaluation}

The study then proceeded to fine-tune the LLaMA 2 model with the prepared training dataset. The process was set for six epochs, with a batch size of four and a learning rate of 1e-4. The training process observed a notable reduction in the evaluation loss on the validation dataset, with the loss progressively declining from 0.3251 to 0.2533. The fine-tuned model was imported then to infer the flight trajectory reconstruction outputs for the 910 trajectories in the testing dataset, using input prompts formatted identically to those in the base model evaluation. The following subsections detailed the outcomes of the fine-tuned model, categorized by different flight trajectory characteristics.

\subsubsection{Linear Flight Trajectories} 

\begin{figure*}[t!]
    \centering
    \begin{subfigure}[htbp]{0.49\textwidth}
        \centering
        \includegraphics[width=0.4\textwidth]{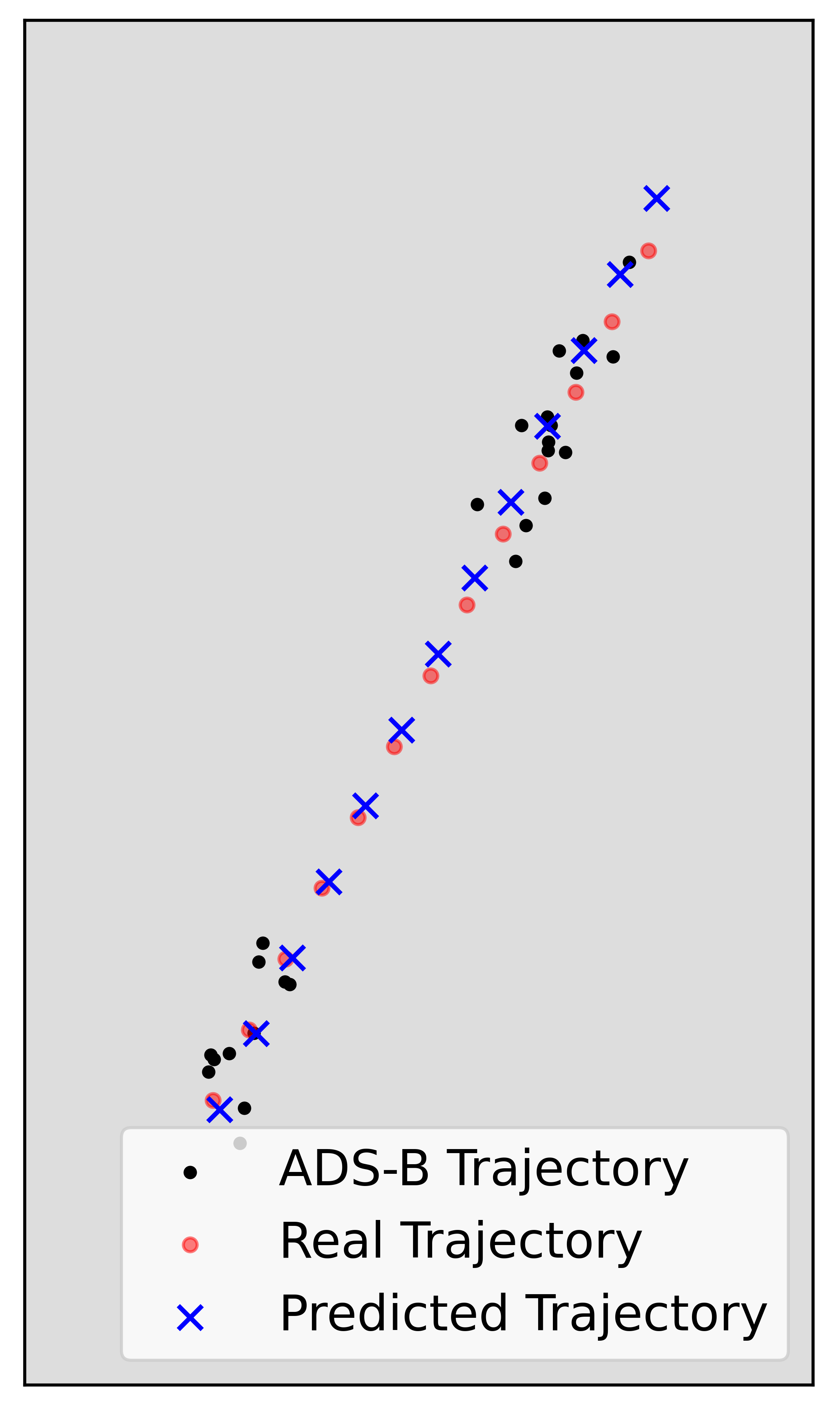}
        \includegraphics[width=\textwidth]{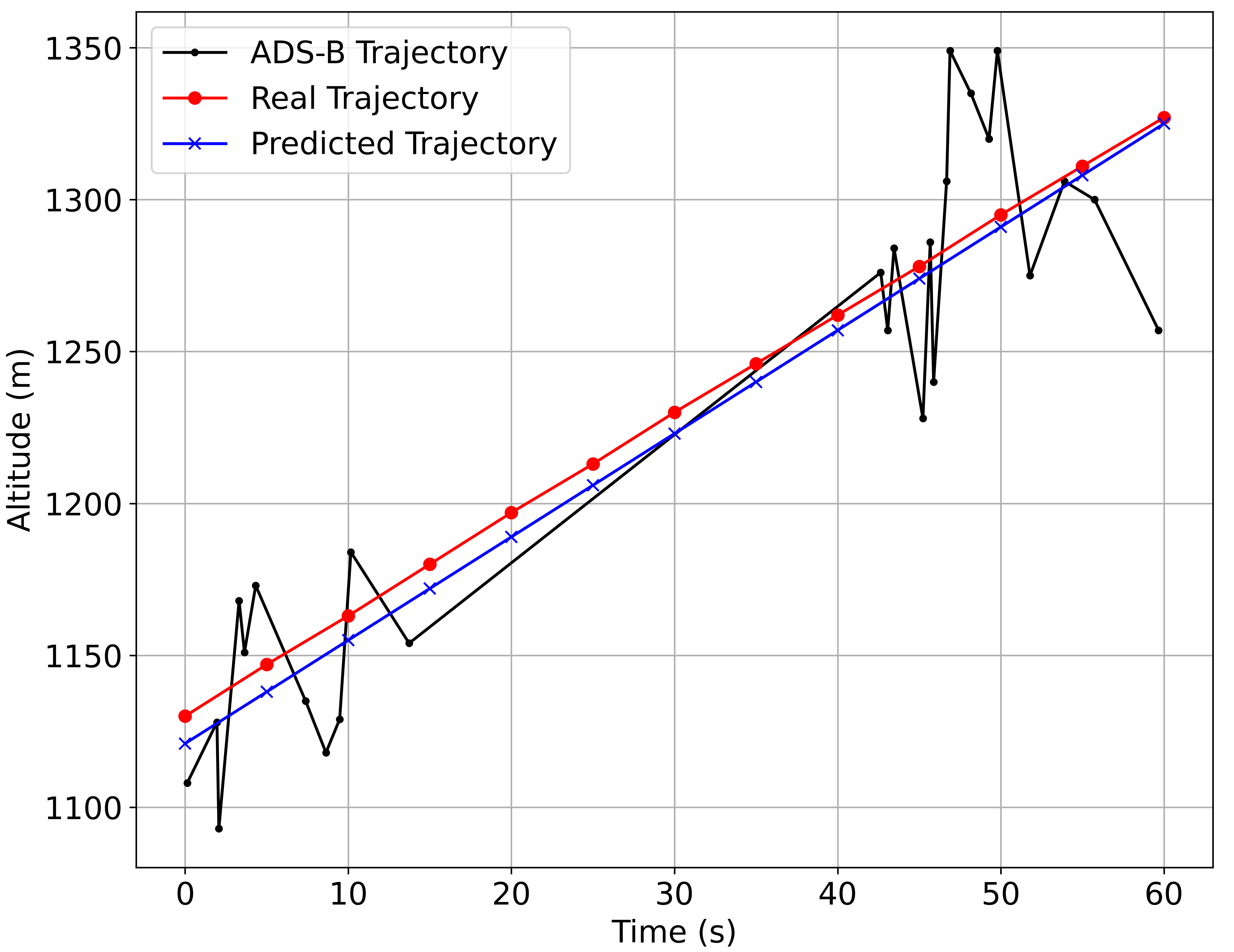}
        \caption{Climbing with constant rate.}\label{fig:trj103}
    \end{subfigure}%
    ~    
    \begin{subfigure}[htbp]{0.49\textwidth}
        \centering
        \includegraphics[width=0.42\textwidth]{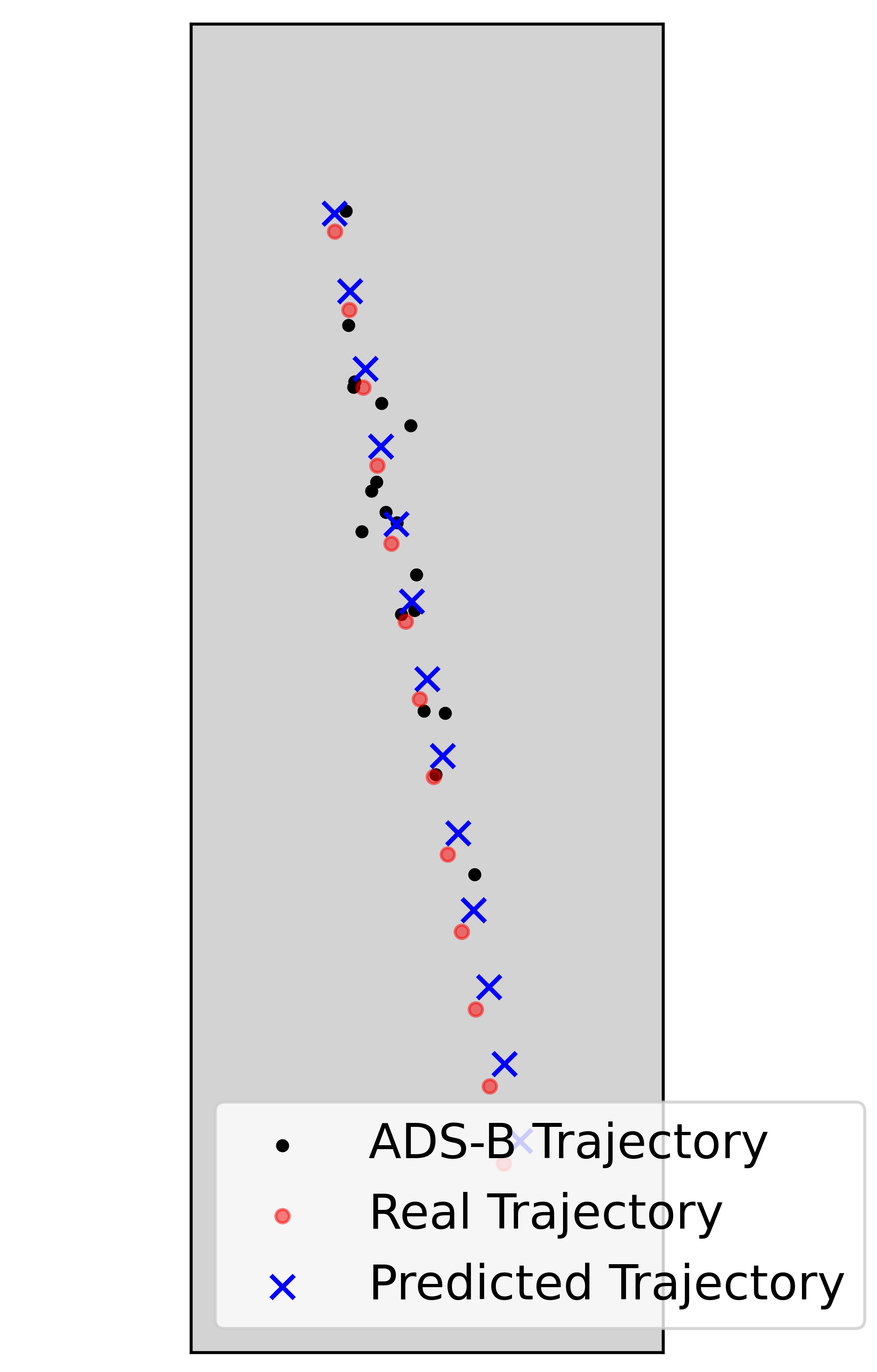}
        \includegraphics[width=\textwidth]{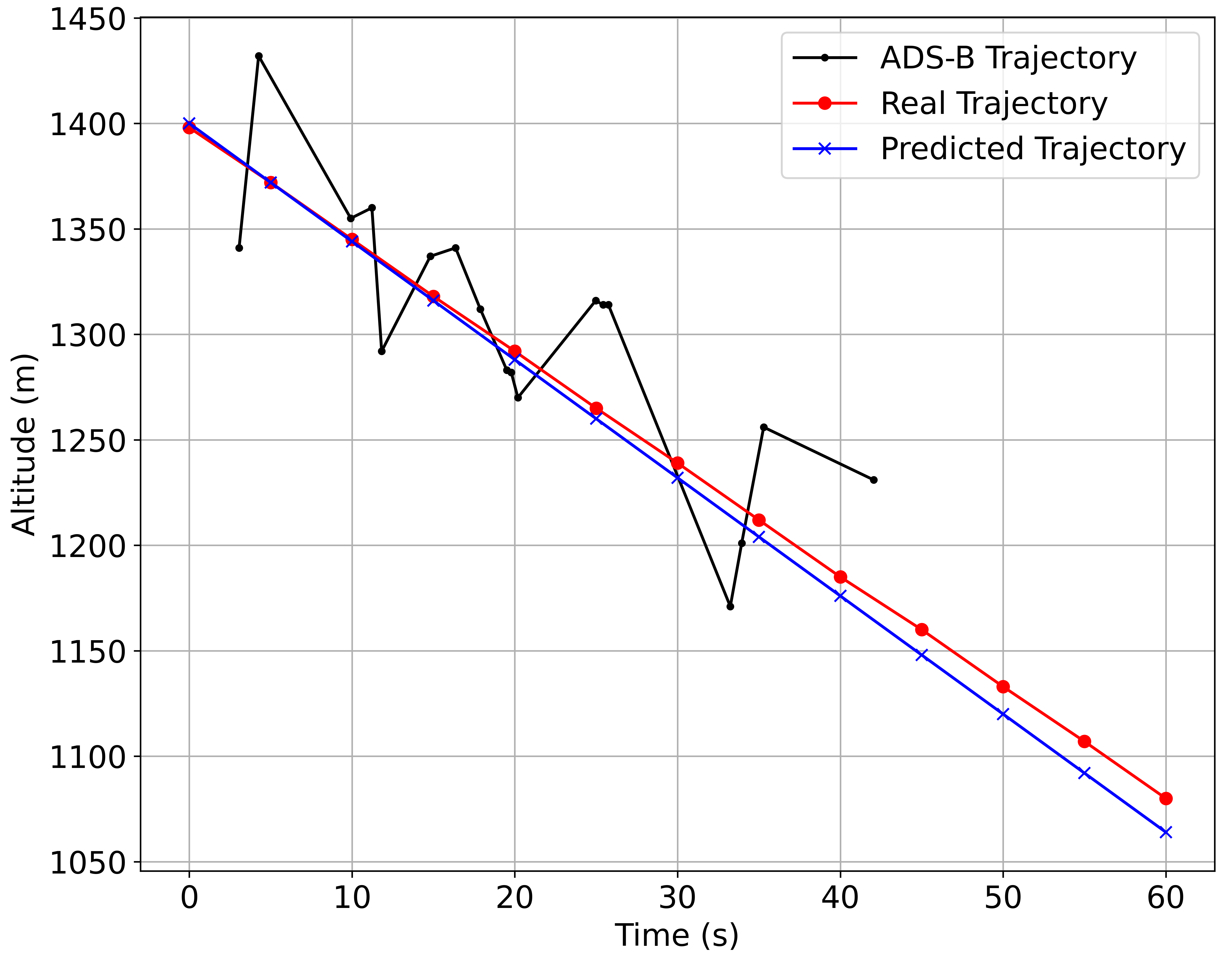}
        \caption{Descending with constant rate.}\label{fig:trj242}
    \end{subfigure}
    \caption{Flight trajectory reconstruction results for linear flight trajectories.}\label{fig:trj-linear}
\end{figure*}

The initial evaluation focused on linear flight trajectories, which are the most commonly observed and relatively simple in nature. Figure~\ref{fig:trj103} and Figure~\ref{fig:trj242} illustrate two scenarios, one depicting a climbing trajectory and the other a descending trajectory, both demonstrating a linear pattern on a 2D plane.

Analysis of Figure~\ref{fig:trj-linear} reveals that the model can successfully execute the task of flight trajectory reconstruction after fine-tuning. The predicted trajectory, marked by blue crosses, aligns closely with the ground truth trajectory, marked by red dots. The black dots scattered in the figure represent the ADS-B data points, which have been subjected to noise manipulation and data omission. Notably, the model effectively mitigated the noise and accurately estimated the aircraft's position at each integer timestamp. The altitude predictions also exhibited considerable accuracy, particularly when the ADS-B data was available. However, some deviations were observed in the absence of ADS-B data, which is reasonable since the model has no reference to infer the altitude, considering the short duration of the flight trajectory.

\subsubsection{Curved Flight Trajectories}
\begin{figure*}[t!]
    \centering
    \begin{subfigure}[t]{0.48\textwidth}
        \centering
        \includegraphics[width=0.5\textwidth]{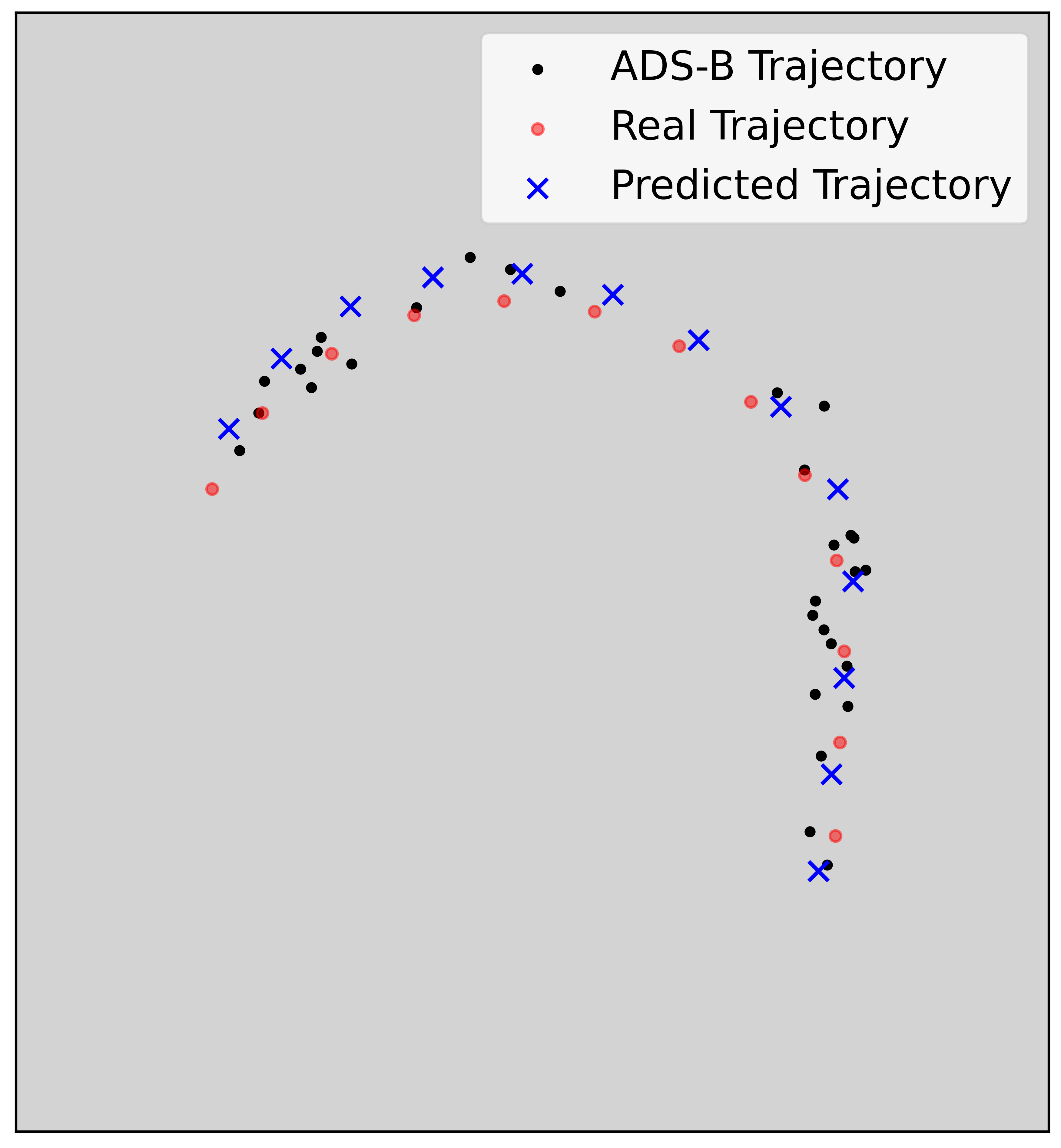}
        \includegraphics[width=\textwidth]{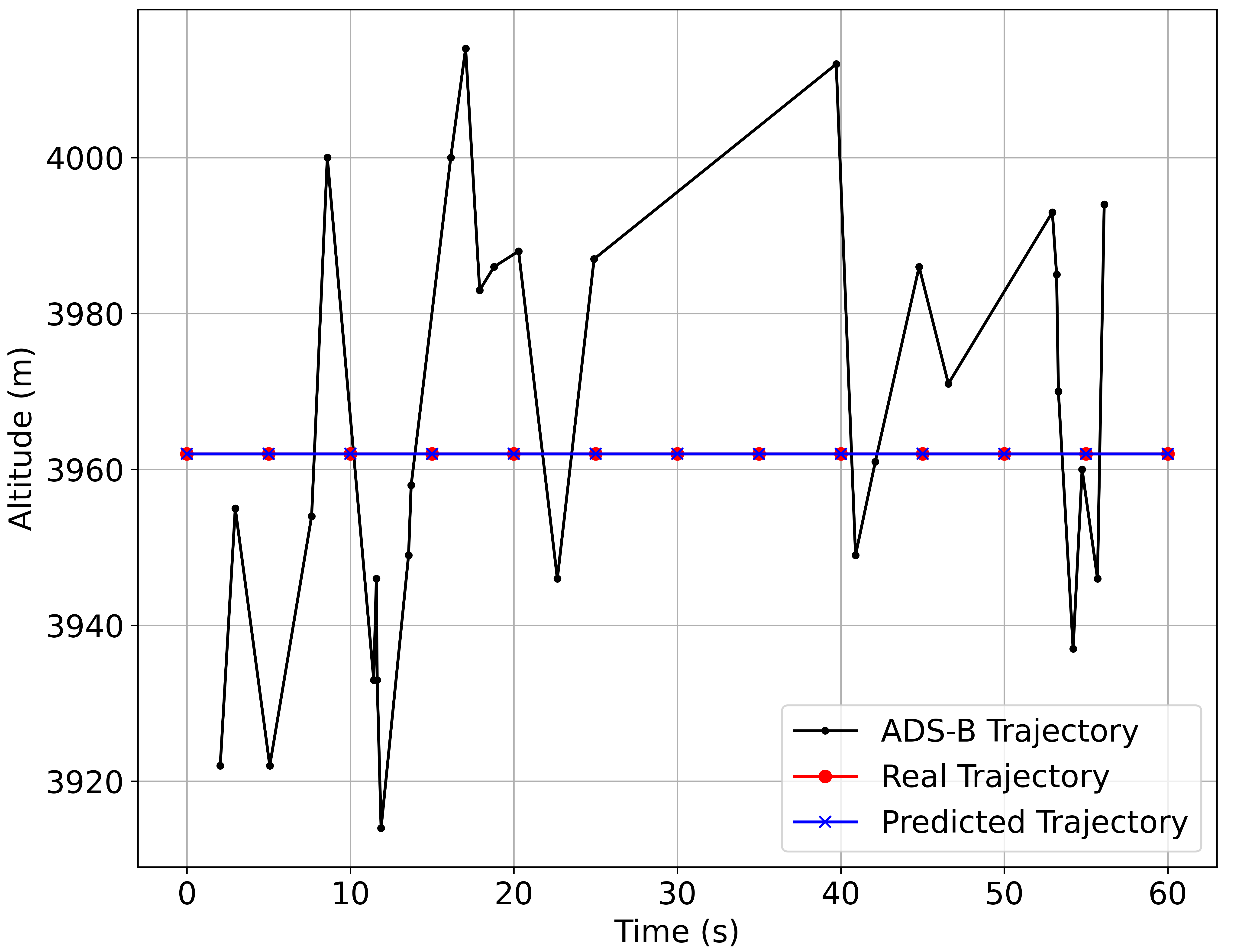}
        \caption{Curved flight trajectory maintaining a level altitude.}\label{fig:trj464}
    \end{subfigure}
    ~
    \begin{subfigure}[t]{0.48\textwidth}
        \centering
        \includegraphics[width=0.5\textwidth]{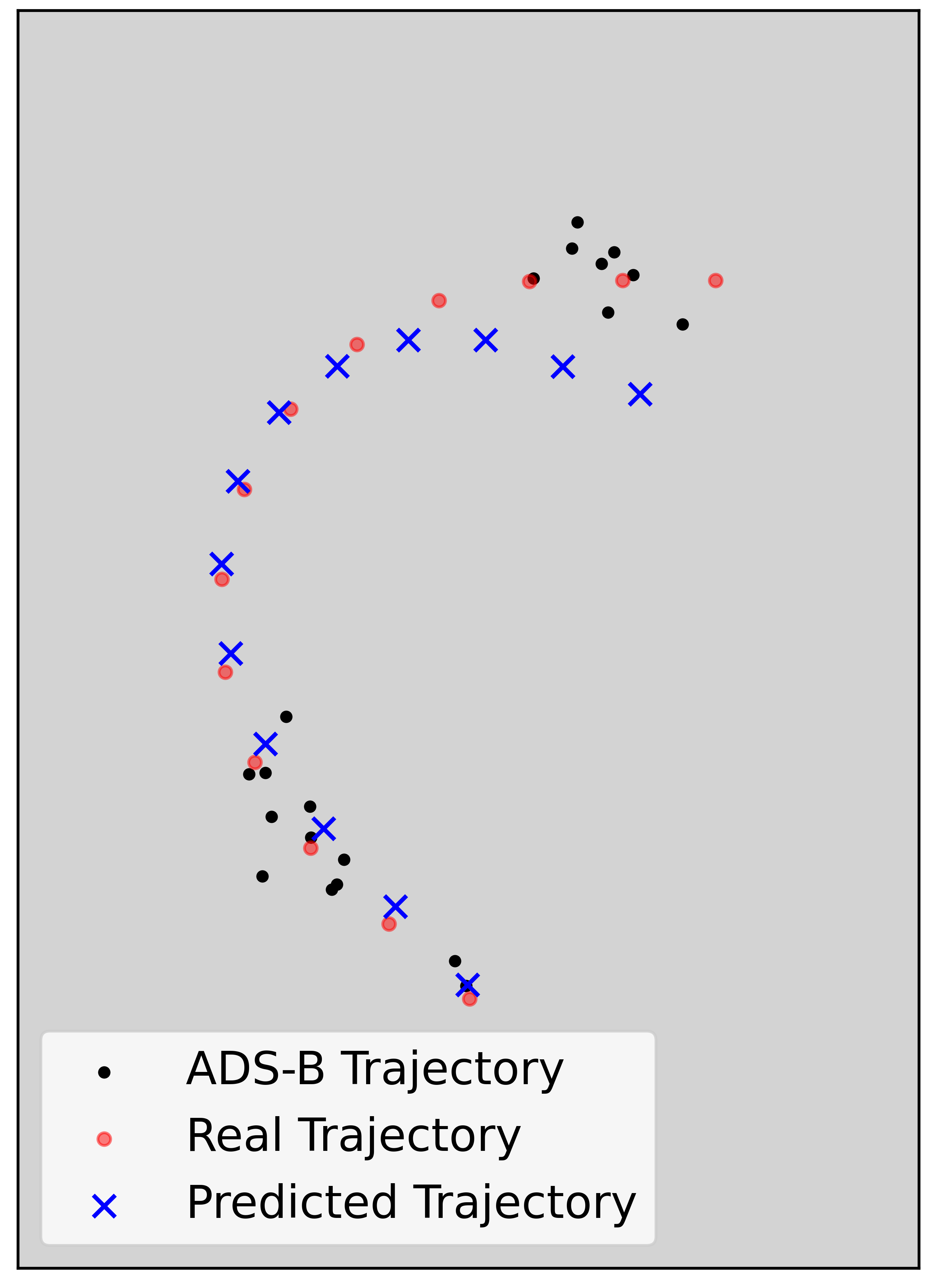}
        \includegraphics[width=\textwidth]{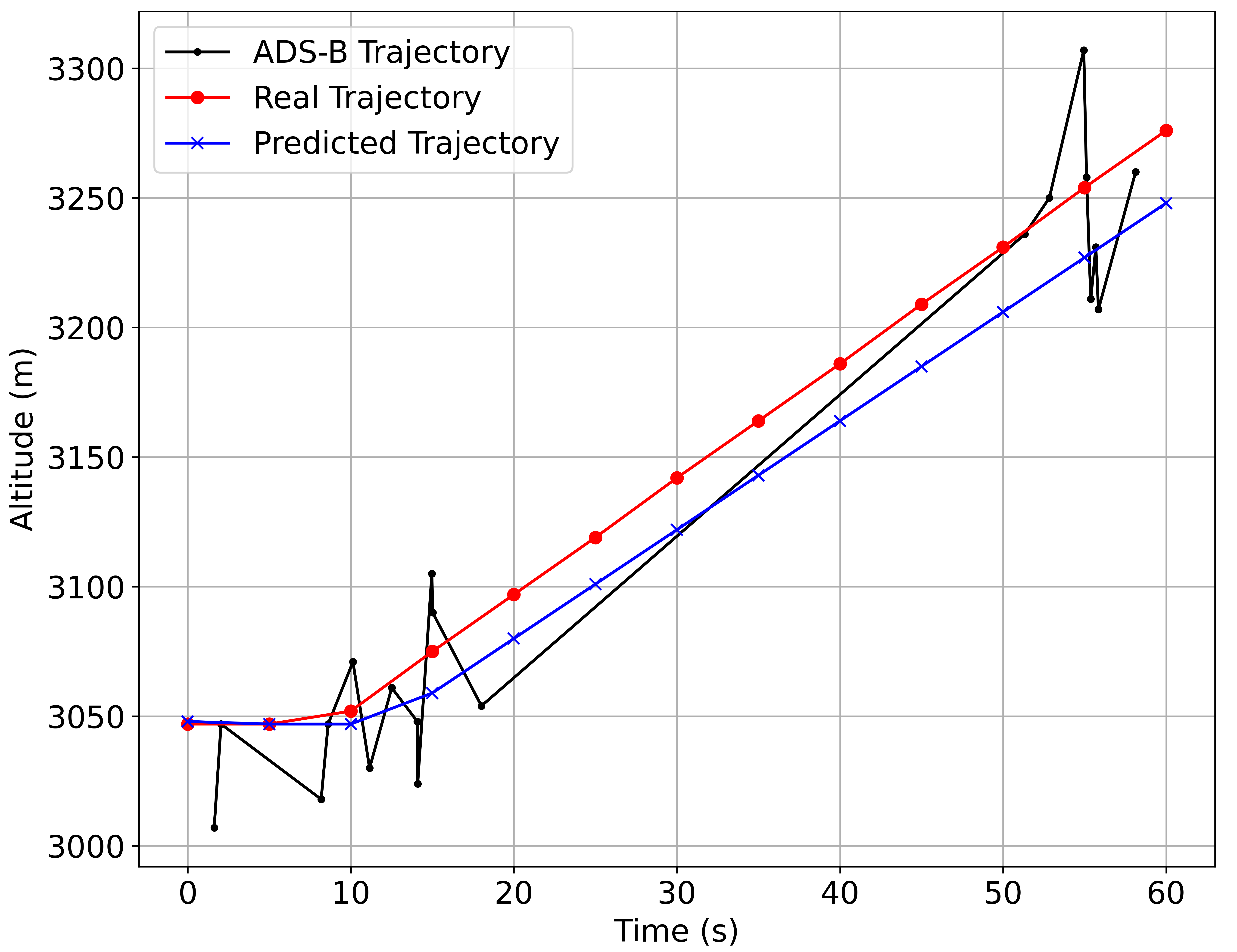}
        \caption{Curved flight trajectory exhibiting a climbing motion.}\label{fig:trj444}
    \end{subfigure}
    \caption{Curved flight trajectory with level and changing altitude.}\label{fig:trj-curved}
\end{figure*}

The study next examined the model's performance with curved flight trajectories, which present a higher level of complexity than linear flight trajectories. Analysis of Figure~\ref{fig:trj-curved} reveals that the model satisfactorily met the task expectations in these scenarios. Particularly, even in cases where the ADS-B data was missing along the curved segments, the model still generated a trajectory aligned with the ground truth trajectory. 

Specifically, Figure~\ref{fig:trj464} highlights the model's proficiency in filtering altitude data noise, accurately reflecting the aircraft's level flight. Moreover, the model also demonstrated the capability to estimate fluctuating altitudes, such as those experienced during turbulent conditions, as evidenced by the non-smooth altitude line in other observations. Conversely, Figure~\ref{fig:trj444} indicates the limitations in the model's prediction accuracy. Specifically, the model's prediction exhibited latency and deviation, particularly in the later stages of the flight trajectory. This latency may be caused by the model's inability to learn the aircraft's behavior due to the short duration of the flight data. However, the altitude graph in Figure~\ref{fig:trj444} shows a commendable alignment of the predicted and ground-truthing climbing rates, despite the delay in the model's prediction of the climbing action.

\section{Discussion}

\subsection{Limitations}

The experiments in the study have successfully demonstrated the LLM's potential to process sequential data and discern patterns in reconstructing flight trajectories with limited training data. However, as the LLMs take the numbers as tokens, a notable limitation is the LLM's constraint on token length, which significantly restricts its practical applicability in more extensive data scenarios. Attempted tests conducted with input duration beyond 60 seconds, such as 90 seconds and 120 seconds, resulted in the token length exceeding the model's recommended length of 2048 tokens, e.g., 3072 or 4096 tokens. The model in those tests showed reasonable accuracy in initial predictions but suffered from increasingly inaccurate predictions as the sequences progressed. This phenomenon suggests a loss of context information retention in a longer text sequence. Additionally, as the token length increases, LLM inference speed decrease is observed, further impacting its practical utility. Therefore, the current LLMs cannot yet replace established methodologies, such as Kalman filtering, in precision and efficiency. Nevertheless, given the rapid advancements in LLM's capabilities, their potential in time-series data processing is promising.

Moreover, the study's scope was limited to the LLaMA 2 - 7B model, without exploring other LLMs, such as GPTs, or training a larger parameter size model. Future studies should aim to conduct a comparative analysis across various LLMs and model sizes, examining the impact of parameter configuration and PEFT methods on prediction accuracy. Additionally, the evaluation metrics in future studies could not be limited to focus on direct distance measures. A more comprehensive evaluation approach could include assessing the probability of predictions falling within a predefined ground truthing cylindrical range.

\subsection{Future Research}

Based on the findings of this study, several avenues for future research emerge. One of the primary goals should be to address the token length limitation of LLMs. Advancements in this area will enable the LLMs to process longer sequences and a diverse range of data, broadening the scope and applicability of LLMs in various fields. For example, the LLMs could be applied to flight phase identification, a classification problem that requires the model to identify the flight phase based on the flight trajectory. This could involve leveraging minimal manually labeled accurate data to train the model, which could have significant implications in aviation management analysis.

Briefly, the potential of LLMs to move towards AGI is an exciting prospect. The ability of these models to learn from diverse data sources and continuously improve their performance suggests that LLMs offer substantial enhancements in various fields. Particularly in the aviation and transportation industry, the application of LLMs could revolutionize data integration and analysis,  predictive modeling, and operational optimization.

\section{Conclusion}
In summary, this study has demonstrated the potential of LLMs in processing time-series data, particularly in flight trajectory reconstruction. Despite the current study's limitations in terms of model capabilities and evaluation scope, it lays the groundwork for future research in this area. Addressing these limitations presents a promising future for advancing the application of LLMs in the aviation and transportation sectors.

\section*{Acknowledgment}

The authors wish to express their gratitude to Dr. John A. Springer for providing essential computing resources and support, which significantly contributed to the research presented in this paper.


\bibliographystyle{IEEEtran}
\bibliography{IEEEabrv,mybib}

\end{document}